\documentclass[nohyperref]{article}

\usepackage{hyperref}

\usepackage{times,url,booktabs,microtype,graphicx,subfigure,bbm,verbatim,amsthm,amssymb,amsmath,enumitem,pifont,makecell,multirow,mdframed,xcolor,colortbl,rotating,wrapfig,resizegather,tcolorbox,resizegather,algorithmic,algorithm}
  
\usepackage{caption}
\newcommand{\gr}{\rowcolor[gray]{.95}}

\usepackage[accepted]{icml2023}

\usepackage{amsmath}
\usepackage{amssymb}
\usepackage{mathtools}
\usepackage{amsthm}
\usepackage[export]{adjustbox}
\usepackage{booktabs, makecell, tabularx}
\usepackage{textpos}
\usepackage{multirow}
\usepackage[capitalize,noabbrev]{cleveref}

\theoremstyle{plain}

\theoremstyle{definition}

\theoremstyle{remark}

\usepackage[textsize=tiny]{todonotes}

\icmltitlerunning{Are Large Kernels Better Teachers than Transformers for ConvNets?}

\begin{document}

\twocolumn[
\icmltitle{Are Large Kernels  Better Teachers than Transformers for ConvNets?}




\begin{icmlauthorlist}
\icmlauthor{Tianjin Huang}{1}
\icmlauthor{Lu Yin}{1}
\icmlauthor{Zhenyu Zhang}{2}
\icmlauthor{Li Shen}{3}
\icmlauthor{Meng Fang}{4,1}\\
\icmlauthor{Mykola Pechenizkiy}{1}
\icmlauthor{Zhangyang Wang}{2}
\icmlauthor{Shiwei Liu}{2}
\end{icmlauthorlist}

\icmlaffiliation{1}{Department of Mathematics and Computer Science, Eindhoven University of Technology}
\icmlaffiliation{2}{Department of Electrical and Computer Engineering, University of Texas at Austin}
\icmlaffiliation{3}{JD Explore Academy}
\icmlaffiliation{4}{Department of Computer Science, University of Liverpool}

\icmlcorrespondingauthor{Tianjin Huang}{t.huang@tue.nl}
\icmlcorrespondingauthor{Shiwei Liu}{shiwei.liu@austin.utexas.edu}

\icmlkeywords{Machine Learning, ICML}

\vskip 0.3in
]



\printAffiliationsAndNotice{}  

\begin{abstract}

This paper reveals a new appeal of the recently emerged large-kernel Convolutional Neural Networks (ConvNets): as the teacher in Knowledge Distillation (KD) for small-kernel ConvNets. While Transformers have led state-of-the-art (SOTA) performance in various fields with ever-larger models and labeled data, small-kernel ConvNets are considered more suitable for resource-limited applications due to the efficient convolution operation and compact weight sharing. KD is widely used to boost the performance of small-kernel ConvNets. However, previous research shows that it is not quite effective to distill knowledge (e.g., global information) from Transformers to small-kernel ConvNets, presumably due to their disparate architectures. We hereby carry out a first-of-its-kind study unveiling that modern large-kernel ConvNets, a compelling competitor to Vision Transformers, are remarkably more effective teachers for small-kernel ConvNets, due to more similar architectures. Our findings are backed up by extensive experiments on both logit-level and feature-level KD ``out of the box",  with no dedicated architectural nor training recipe modifications. Notably, we obtain the \textbf{best-ever pure ConvNet} under 30M parameters with \textbf{83.1\%} top-1 accuracy on ImageNet, outperforming current SOTA methods including ConvNeXt V2 and Swin V2. We also find that beneficial characteristics of large-kernel ConvNets, e.g., larger effective receptive fields, can be seamlessly transferred to students through this large-to-small kernel distillation. Code is available at: \url{https://github.com/VITA-Group/SLaK}.


\end{abstract}
\section{Introduction}
\label{sec:intro}
Transformers~\cite{vaswani2017attention} have shined in the past two years, bringing a historical revolution in artificial intelligence, from the foundation models in natural language processing~\cite{brown2020language,ramesh2022hierarchical,du2022glam,li2023cancergpt,chen2023sparse,chowdhery2022palm} to the Vision Transformers in computer vision~\cite{dosoViTskiy2021an,liu2021swin}, to biological sciences~\cite{jumper2021highly}, etc. By leveraging larger and larger models and labeled data, Transformers continuously establish new state-of-the-art (SOTA) performance bars in various fields. Despite their impressive performance, Transformers arguably cause a prohibitive competition of gigantic models in academia and industry, pushing the SOTA model size beyond the reach of common hardware.


On the other hand, Convolutional Neural Networks (ConvNets) still hold their merits in computer vision scenarios with limited computational resources, such as Edge AI~\cite{li2019edge} and artificial intelligence of things (AIoT)~\cite{zhang2020empowering}. Compared to advanced Vision Transformers, small-kernel ConvNets like ResNets~\cite{resnet} and MobileNets~\cite{howard2017mobilenets} are generally less expensive to train and infer, despite typically having lower performance. Therefore, improving the performance of small-kernel ConvNets to the state-of-the-art levels achieved by advanced Vision Transformers while maintaining model size affordable has excellent value in real-world environments. However, this is a non-trivial research question since previous work~\cite{yao2023distill} has shown that distilling knowledge from Vision Transformers to small-kernel ConvNets can be ineffective or even detrimental. In this study, we carry out a first-of-its-kind study unveiling that modern large-kernel ConvNets, a compelling competitor to Vision Transformers, are more effective, and easy-to-adopt teachers for small-kernel ConvNets due to their architectural similarities.

Knowledge Distillation (KD)~\cite{hinton2015distilling} with its variants~\cite{yuan2020revisiting,zhou2021rethinking,zhao2022decoupled} has evolved as one of the leading approaches to enhancing the performance of deep neural networks. The basic concept is to utilize larger teacher models to generate pseudo-labels for smaller models, which are subsequently trained to mimic the teacher's behavior, without increasing the model size. While distilling knowledge from advanced Vision Transformers to small-kernel ConvNets seems to be a logical choice to boost the performance of the latter, doing so has little efficacy likely due to the inherent architectural discrepancy in between~\cite{yao2023distill}.

Recently emerged \textbf{large-kernel ConvNets}, e.g., ConvNeXt~\cite{liu2022convnet} and RepLKNet~\cite{ding2022scaling}, demonstrate that pure convolutional models can deliver the same excellence as Vision Transformers when equipped with similar advanced designs, whose performance can be further stretched out by purely enlarging kernel size to 51$\times$51 in SLaK~\cite{liu2022more}. Yet, the effectiveness of large kernel ConvNets in teaching compact models remains to lack intuition and is unexplored. 

Intuitively, large-kernel ConvNets have three advantages (which can be satisfied simultaneously or partially) compared to Vision Transformers, as teachers for small-kernel ConvNets: (1) equally good accuracy; (2) similar or even larger effective receptive field (ERF); (3) more importantly, convolutional operations instead of self-attention modules. However, these modern ConvNets (e.g., ConvNexT and SLaK) also share many discrepancies with conventional small-kernel ConvNets, including but not limited to ``patchify stem'', inverted bottleneck, BatchNorm instead of LayerNorm, and replacing ReLU with GELU~\cite{liu2022convnet}. Hence it remains unclear whether large-kernel ConvNets can perform better than Vision Transformers or not when distilled to small-kernel ConvNets.

In this paper, we conduct a systematic comparison between modern large-kernel ConvNets (such as ConvNeXt and SLaK) and advanced Vision Transformers (such as ViT~\cite{dosoViTskiy2021an}, Swin~\cite{liu2021swin}, and CSWin~\cite{dong2022CSWin}) as teacher models, when distilled into small-kernel ConvNet student. We discover that large-kernel ConvNets are significantly more effective teachers than  Vision Transformers for small-kernel ConvNets, for both feature-level and logit-level KD approaches.   We also find that our distilled models enjoy larger ERF and better robustness than others, indicating that, besides accuracy, other good properties of large kernels can also be seamlessly transferred to small kernels through our large-to-small kernel distillation paradigm.  Our contribution can be summarized as follows:
\vspace{-0.5em}


\begin{itemize}
    \item We conduct a pioneering empirical study of large-to-small kernel distillation. Through a thorough comparison of advanced Vision Transformers (including ViT, Swin, and CSWin) and modern large-kernel ConvNets (ConvNeXt and SLaK) when distilled to small-kernel ConvNets, we discover two principles for ConvNet distillation: \ding{182} large-kernel ConvNets function as more effective teachers than Transformers for small-kernel ConvNets; \ding{183} among large-kernel teachers, students obtain greater benefits from larger kernels compared with the smaller ones.  
    \item Following the above principles without any dedicated architectural and training designs, we favorably distill the recently proposed 51$\times$51 SLaK-T into 7$\times$7 ConvNeXt-T, obtaining the best-ever 30M ConvNet with \textbf{83.1\%} top-1 accuracy on ImageNet, outperforming the current state-of-the-arts ConvNeXt V2 and Swin V2. Interestingly, the distilled ConvNeXt-T even outperforms the strong result of its teacher by 0.6\%.
    \item More interestingly, we  found that students distilled from larger kernels are automatically embedded with better robustness, as well as larger and denser ERF than the ones distilled from small kernels or Transformers. Given the recently emerging trend that correlates larger ERF with better performance~\cite{ding2022scaling,kim2021dead,liu2022more,dai2022demystify,yang2022fast}, this observation also explains the success of large-to-small kernel distillation. 

\end{itemize}

\section{Related Work}
\subsection{Knowledge Distillation}
\looseness=-1  Knowledge distillation (KD) is proposed by~\citet{hinton2015distilling}, to transfer knowledge from one teacher model to another student model, by forcing the latter to mimic the prediction of the teacher.  
Since then, various variants of knowledge distillation have been proposed \cite{
fang2022up, fang2021mosaicking, xue2021kdexplainer}, which can be roughly categorized into two groups: \underline{1) Distillation from logits}~\cite{cho2019efficacy,yang2019snapshot,mirzadeh2020improved,yang2022rethinking,yao2023distill}. Logit-level distillation mainly try to minimize the KL-Divergence between prediction logits of teachers and students. Recently,~\citet{zhao2022decoupled} decouple the classical KD loss into two parts, i.e., target class knowledge distillation and non-target class knowledge distillation, achieving better results.~\citet{yang2022rethinking} revise the formulation of classical KD and found that besides cross-entropy (CE) loss, it also contains an extra loss which contains the knowledge of all classes except the target class. By adding teachers’ target output as an extra soft loss, they can outperform the  previous logit distillation approaches, like KD, DKD~\cite{yang2022focal}, etc. 
\underline{2) Distillation from representation}~\cite{heo2019comprehensive,heo2019knowledge,yang2021knowledge,huang2017like,kim2018paraphrasing,park2019relational,yim2017gift,zagoruyko2016paying,wei2022contrastive}.~\citet{romero2014fitnets} distill the learned knowledge from the intermediate feature directly.~\citet{park2019relational} extract relations from the feature map and transfer the relation instead.~\citet{chen2021distilling} distill knowledge from multi-level features map. Recently,~\citet{yang2022masked} proposes to let the student model generate the teacher model's feature instead of mimicking (MGD), improving the effectiveness of feature distillation. \citet{yang2022vitkd} further comprehensively studies the effect of mimicking KD and Generation KD on ViT distillation, resulting in a suite of principles for feature-based ViT distillation.

\citet{li2022spatial} introduce spatial-channel token distillation to mix up the information in both the spatial and channel dimensions for Vision MLP. Multiple receptive tokens are applied in dense prediction tasks to indicate the pixels of interest in the feature map, with a distillation mask generated via pixel-wise attention~\citep{huang2022masked}.~\citet{huang2022knowledge} revisit the fact the training with much stronger teachers could significantly hurt student's performance~\cite{cho2019efficacy} and address it by simply preserving the relations between the prediction of teacher and student with a correlation-based loss. Very recent work~\cite{yao2023distill} unveils that distilling knowledge (e.g., global information) from Vision Transformers to ConvNets is ineffective likely due to their architectural gaps. In this work, instead of distilling knowledge from Transformers, we introduce large-to-small kernel distillation, which can seamlessly transfer preferable knowledge (e.g., accuracy, global information, and robustness) to small-kernel ConvNets.

\subsection{Large Kernel Convolutions}
Large kernel convolutions date back to the 2010s, where AlexNet~\cite{krizhevsky2012imagenet} adopts 11$\times$11 kernels in the first convolutional layer and Inception~\cite{szegedy2015going,szegedy2017inception} stacks a combination of 1$\times$7 and 7$\times$1 kernels. Global Convolutional Network~\cite{peng2017large} constructs ConvNets with a pair of stacked large convolution, with kernel size up to 25$\times$1 $+$ 1$\times$25. Since the popularity of VGG~\cite{simonyan2014very}, people favor small stacked 3$\times$3 kernels to build ConvNets~\cite{resnet,howard2017mobilenets,xie2017aggregated,densenet}. Although computationally efficient, it is not very efficient for small kernels to achieve large effective receptive fields (ERF), even with over 100 layers in the model. 

Inspired by the local window (at least 7$\times7$) self-attention used in Swin Transformers~\cite{liu2021swin},~\citet{liu2022convnet} revisit the use of large kernel-sized convolutions for ConvNets, finding that 7$\times$7 depthwise convolutions consistently outperform 3$\times$3 in ConvNeXt. 
RepLKNet~\cite{ding2022scaling} continues to scale kernel size to 31$\times$31 using  Structural Reparameterization while achieving comparable or superior results than Swin Transformer. SLaK~\cite{liu2022more} pushes along the direction of pure ConvNets with extremely large kernels up to 51$\times$51, which has never been discussed before. More recently,~\citet{chen2022scaling,xiao2022dynamic} reveals the feasibility of large kernels for 3D ConvNets and time series classification, respectively. SegNeXt~\cite{guo2022segnext} ensembles a novel convolutional attention network with multiple large kernels, showing strong results on semantic segmentation. Our paper differs from the previous work and focuses on an empirical pilot study of knowledge distillation from large kernels to small kernels. 

\vspace{-0.5em}
\section{Experimental Setup}
\label{sec:exp_setup}
In this section, we describe the experimental setup and benchmarks used. Our goal is to comprehensively compare Vision Transformers and modern large-kernel ConvNets in the context of knowledge distillation and to study which is more suitable as teachers for small-kernel ConvNets. To enable fair comparisons, we carefully investigate several key components and summarize them below.

\textbf{Evaluation Metrics.} Given a teacher model (T) with high accuracy on a task $acc\,(teacher)$ and a student model (S) with lower accuracy $acc\,(student)$, we can improve the accuracy of the latter to $acc\, (distilled)$ by via knowledge distillation. To enable comparisons among different teacher models, we choose two metrics to report, \underline{Direct Gain} and \underline{Effective Gain}. Direct Gain refers to the direct performance difference with and without knowledge distillation, i.e.,
\begin{align}
  \text{Direct Gain} = acc\, (distilled) - acc\,(student)
\end{align}
Ideally, we hope all the teachers have the same accuracy to make a fair comparison solely for their distillation effectiveness. However, different models inevitably have accuracy discrepancies, impacting our comparisons. To mitigate this undesirable effect, we scale the Direct Gain by teachers' accuracy, resulting in Effective Gain:
\begin{align}
  \text{Effective Gain} =\! \frac{acc\, (distilled) - acc\,(student)}{acc\,(teacher)}
\end{align}
In all our experiments, we report these two metrics.

\textbf{Dataset, Teacher and Student Models.} 
We conduct experiments on the commonly used ImageNet-1K dataset~\cite{russakovsky2015imagenet} containing 1k classes, 1,281,167 training images, and 50,000 validation images. 

We have two main distillation pipelines: \underline{Pipeline I}: Large-kernel ConvNets to small-kernel ConvNets distillation; \underline{Pipeline II}: Transformers to small-kernel ConvNets distillation. 
For both pipelines, we opt for two ConvNets as our students: ResNet-50 with $3\times3$ kernels as the most representative conventional ConvNet, and ConvNeXt-T\footnote{Follow~\cite{liu2022more}, we add a BatchNorm layer after $7\times7$ kernels since it gives us consistently better accuracy.} with $7\times7$ kernels as the most representative modern ConvNet. ConvNeXt-T also serves suitable small-kernel ConvNet when it comes to teachers like SLaK-T ($51\times51$ kernels), ViT (global attention), Swin-T (at least $7\times7$ windows), and CSWin-T (global cross-shaped windows). Our teacher models are ConvNeXt-T and SLaK for Pipeline I; ViT-S, Swin-T, and CSWin-T for Pipeline II. Overall, the teachers in Pipeline I and Pipeline II are of similar sizes and comparable accuracy, ensuring a fair comparison. The pre-trained weights of SLaK-T, ConvNeXt-T, and CSWin-T are downloaded from their official GitHub repositories. The pre-trained weights of ViT-S and Swin-T are downloaded from TIMM~\cite{rw2019timm}. 

\textbf{Training Recipe.} In our study, we use a set of training recipes closely following DeiT~\cite{touvron2021training}, Swin Transformers~\cite{liu2021swin}, and ConvNeXt~\cite{liu2022convnet}. We use AdamW optimizer \citep{loshchilov2018decoupled} and train models for  120 epochs (Section~\ref{sec:LK_vs_Trans}) and 300 epochs (Section~\ref{sec:300-epochs})  with a batch size of 4096, and a weight decay of 0.05. The learning rate is 4e-3 with a 20-epoch linear warmup followed by a cosine decaying schedule. 

For data augmentation, we use the default RandAugment \citep{cubuk2020randaugment} in Timm \citep{rw2019timm} -- ``rand-m9-mstd0.5-inc1'', Label Smoothing \citep{szegedy2016rethinking} coefficient of 0.1, Mixup \citep{zhang2017mixup} with $\alpha = 0.8$, Cutmix \citep{yun2019cutmix} with $\alpha = 1.0$, Random Erasing \citep{zhong2020random} with $p=0.25$, Stochastic Depth with a drop rate of 0.1 for ConvNeXt-T and 0.0 for ResNet-50, and Layer Scale \citep{touvron2021going} of the initial value of 1e-6. We train all models with 4 NVIDIA A100 GPUs.

\begin{table*}[h]
	\caption{\textbf{NKD~\cite{yang2022rethinking} results of ResNet-50 and ConvNeXt-T distilled from various teachers on ImageNet.}  All models are distilled for 120 epochs.  Effective Gain is defined as $\frac{acc\, (distilled) - acc\,(student)}{acc\,(teacher)}$, and Direct Gain is $acc\, (distilled) - acc\,(student)$. }
	\label{tab:distill_logits_120_NKD}
	\begin{center}
  \resizebox{0.95\textwidth}{!}{
		\begin{tabular}{lcc|c|ccccc}
			\toprule
		  Teacher & Arch. Type & Kernel-Size&Student & Teacher Top-1 &  Student Top-1  & Distilled Top-1 & Effective Gain  & Direct Gain  \\
            \midrule
			ViT-S & Transformer &N/A& ResNet-50 & 79.8 & 76.13 & 77.14  & 1.20  & 1.01 \\
		    \midrule
      	Swin-T & Transformer &N/A&  ResNet-50 & 81.3 & 76.13 & 77.67 & 1.89 &  1.54 \\
			\midrule
            CSWin-T & Transformer &N/A&  ResNet-50 &82.7 &76.13  & 77.68 & 1.87 & 1.55 \\
            \midrule
            ResNet-50&ConvNet&3$\times$3&ResNet-50&80.4&76.13&77.82&2.1&1.69\\
            \midrule
		    ConvNeXt-T & ConvNet &7$\times$7& ResNet-50 &  82.1  & 
            76.13 & 78.04 & 2.30  & 1.91 \\
			\midrule
            SLaK-T  & ConvNet &51$\times$51& ResNet-50 & 82.5  & 76.13 & \textbf{78.57} & \textbf{2.90}  & \textbf{2.44} \\
			\midrule
            \midrule
		    Swin-T & Transformer &N/A& ConvNeXt-T &81.3 &81.00  &81.10 &0.12 & 0.10 \\
            \midrule
            CSWin-T & Transformer &N/A& ConvNeXt-T & 82.7 &81.00  & 81.65 & 0.70 & 0.65\\
            \midrule
            ResNet-50&ConvNet&3$\times$3&ConvNeXt-T&80.4&81.00&81.15&0.19&0.15\\
            \midrule
            ConvNeXt-T & ConvNet &7$\times$7 & ConvNeXt-T & 82.1   &81.00  &81.77 &  0.93  & 0.77 \\
            \midrule
            SLaK-T & ConvNet &51$\times$51 & ConvNeXt-T & 82.5  &81.00 &\textbf{82.17}  &  \textbf{1.42} & \textbf{1.17} \\
            \bottomrule
		\end{tabular}}		
	\end{center}
		\vspace{-0.1in}
\end{table*}

\begin{table*}[h]
	\caption{\textbf{KD~\cite{hinton2015distilling} results of ResNet-50 and ConvNeXt-T distilled from various teachers on ImageNet.} All models are distilled for 120 epochs. Effective Gain is defined as $\frac{acc\, (distilled) - acc\,(student)}{acc\,(teacher)}$, and Direct Gain is  $acc\, (distilled) - acc\,(student)$. }
	\label{tab:distill_120_KD}
	\begin{center}
		\footnotesize
    \resizebox{0.95\textwidth}{!}{
		\begin{tabular}{lcc|c|ccccc}
			\toprule
		  Teacher & Arch. Type &Kernel-Size& Student & Teacher Top-1 &  Student Top-1  & Distilled Top-1 & Effective Gain  & Direct Gain  \\
            \midrule
            ViT-S & Transformer &N/A& ResNet-50 &79.8  & 76.13 &76.96  &1.04 & 0.83 \\ 
            \midrule
		    Swin-T & Transformer&N/A & ResNet-50 & 81.3 &76.13  &76.87  & 0.91 &  0.74 \\ 
            \midrule
            CSWin-T & Transformer &N/A&  ResNet-50 &82.7 &76.13  &76.77  & 0.77 & 0.64 \\
      	\midrule
   		ConvNeXt-T & ConvNet &7$\times$7& ResNet-50 &82.1  &76.13 
             &76.93 & 0.97  & 0.80 \\ 
			\midrule
            SLaK-T & ConvNet &51$\times$51& ResNet-50 & 82.5 &76.13 &\textbf{77.05}  & \textbf{1.12}  & \textbf{0.92}  \\ 
            \midrule
            \midrule
		    Swin-T & Transformer &N/A & ConvNeXt-T &81.3 &81.00  & 80.93& -0.08  & -0.07  \\ 
            \midrule
            CSWin-T & Transformer &N/A& ConvNeXt-T & 82.7 &81.00  & 81.18 & 0.22 & 0.18\\
            \midrule
      	ConvNeXt-T & ConvNet &7$\times$7 & ConvNeXt-T &82.1 &81.00 
             & 81.64&0.77  & 0.64  \\ 
			\midrule
            SLaK-T  & ConvNet &51$\times$51& ConvNeXt-T &82.5 &81.00 &\textbf{81.86} &\textbf{1.04}  & \textbf{0.86} \\ 
            \bottomrule
		\end{tabular}}		
	\end{center}
		\vspace{-0.1in}
\end{table*}

\textbf{Distillation Methods.} To draw solid conclusions, we adopt both logit-level distillation and feature-level distillation in this study. Without loss of generality, we opt for the  widely-used Knowledge Distillation (KD)~\cite{hinton2015distilling} as our logit-level method. Additionally, the recently proposed New Knowledge Distillation (NKD)~\cite{yang2022rethinking} is also adopted given its strong results over KD and DKD~\cite{yang2022focal}. We adopt FD~\cite{wei2022contrastive} as our feature-level distillation due to its superior performance. For clarity, we provide the following formal definitions for each method. Let $Z_{t}, Z_{s}$ be the logits of the teacher model and the student model, respectively; $KL(\cdot)$, $\mathcal{L}_{CE}(\cdot)$, and $\phi(\cdot)$ is the Kullback-Leiler divergence loss, the cross-entropy loss, and the softmax function, respectively. We denote $\tau$ as the temperature hyperparameter of KD, $C$ as the classes, (x,y) as the inputs, and $\lambda$ as the coefficient.  

$\bullet$  \textit{Knowledge Distillation (KD)}~\cite{hinton2015distilling}. It consists of the KL divergence loss and the cross-entropy loss. Concretely, the cross-entropy loss encourages the student model to learn knowledge from the true label and the KL divergence transfers the knowledge of the teacher to the student. Formally, it is expressed as follows:  
\begin{align}
    \!\mathcal{L}_{KD}\!=\!(1\!-\!\lambda)\mathcal{L}_{CE}(\phi(Z_s) , y)\!\!+\!\! \lambda \tau^2 KL(\phi(Z_{s}/\tau),\!\phi(Z_{t}/\tau))\! 
\end{align}

$\bullet$  \textit{New Knowledge Distillation (NKD)}~\cite{yang2022rethinking}. Inspired by the cross-entropy loss where the sum of the two distributions are equal, NKD normalizes both the non-target student and teacher output probability such that they meet this constraint as well. Besides, NKD also proposes the soft loss that regards the teacher's target output as the soft target directly. Therefore, NKD consists of the original loss, the non-target distributed loss, and the target soft loss. Formally it is expressed as follows: 
\begin{align}
\mathcal{L}_{NKD}= &-log(\phi(Z_{s}^y))-\phi(Z_{t})^y log(\phi(Z_{s})^y)  \notag\\
                    &- \lambda \tau^2 \sum_{i\neq y}^{C} \widehat{T}^i log(\widehat{S}^i) 
\end{align}
where $\widehat{T}^i=\frac{\phi(Z_{t}/\tau)^i}{1-\phi(Z_{t}/\tau)^y}, \widehat{S}^i=\frac{\phi(Z_{s}/\tau)^i}{1-\phi(Z_{s}/\tau)^y}$ refer to the normalized non-target knowledge and the normalized student's non-target output probability respectively.

$\bullet$ \textit{Feature Distillation (FD)}~\cite{wei2022contrastive}. Feature distillation distills the learned knowledge from intermediate feature maps. Following~\cite{wei2022contrastive}, we apply a 1$\times$1 convolution layer on the top of the student model to align the feature map dimensions of the student model to the teacher model and normalize the output feature map of the teacher model by a whitening operation, implemented by a non-parametric layer normalization operator without scaling and bias.  In distillation, we adopt a smooth $l_{1}$ loss between the student and teacher feature maps, which is formally expressed as follows:
\begin{small}
\begin{align}    
    &\mathcal{L}_{FD}=\mathcal{L}_{CE}(\phi(Z_{s}),y)+\\
                     &\left\{\!
                            \begin{array}{lr}
                            \sum_{l=1}^{L}{\frac{1}{2}(g(S^l)\! -\!norm(T^{l}))^2/\beta}, \|g(S^l)\!-\!norm(T^{l})\|_{1}\! \leq \! \beta, \notag\\
                               \sum_{l=1}^{L}{\|g(S^l)-norm(T^{l})-0.5\beta\|_{1}},\; otherwise\\
                            \end{array}
                      \right.
\end{align}
\end{small}

where $L$ denotes the total number of layers that are used for feature distillation and $S^l$, $T^l$ denotes the $l$-th feature map of the student and the teacher, respectively. $g$ is a 1$\times$1 convolution layer and $norm$ is a non-parametric layer normalization for the whitening operation. $\beta$ is a hyperparameter and is set to 2.0 by default following ~\cite{wei2022contrastive}.



\section{Experimental Results}
\looseness=-1 We report our main results in this section, by evaluating the effectiveness of large-kernel teachers in both logits-level and feature-level KD on ImageNet~\cite{russakovsky2015imagenet}.

\subsection{Large-Kernel ConvNet vs. Transformer as Teachers}
\label{sec:LK_vs_Trans}
\begin{table*}[h]
	\caption{\textbf{Feature distillation results of ResNet-50 distilled from various teacher models on ImageNet dataset.}  The hyper-parameter $L$ is set to 1. Effective Gain is defined as $\frac{acc\, (distilled) - acc\,(student)}{acc\,(teacher)}$, and Direct Gain is $acc\, (distilled) - acc\,(student)$.}
 \vspace{-0.5em}
	\label{tab:distill_FD}
	\begin{center}
   \resizebox{0.95\textwidth}{!}{
		\footnotesize
		\begin{tabular}{lc|ccccccc}
		\toprule
		Teacher & Arch. Type&Student & Teacher Top-1 &  Student Top-1  & Distilled Top-1 & Effective Gain &Direct Gain \\
        \midrule
        \multicolumn{8}{c}{Feature Distillation} \\
        \midrule
        ViT-S & Transformer&ResNet-50 &79.8  & 76.13 &76.81  &0.85 &0.68\\
        \midrule
        Swin-T &Transformer& ResNet-50 & 81.3 &76.13  &76.77 &0.78 & 0.64\\
        \midrule
        ConvNeXt-T &ConvNet& ResNet-50 &82.1  &76.13 
         &76.73 & 0.61&0.70\\
        \midrule
        SLaK-T & ConvNet&ResNet-50 & 82.5 &76.13 &\textbf{76.85}& \textbf{0.87}&\textbf{0.72} \\
        \midrule
        \multicolumn{8}{c}{Feature  + Logits (NKD) Distillation } \\
        \midrule
        ViT-S & Transformer&ResNet-50 &79.8  & 76.13 &76.84  &0.89 &0.71\\
        \midrule
         Swin-T & Transformer&ResNet-50 & 81.3 &76.13  &77.33 &1.47 & 1.20\\
        \midrule
         ConvNeXt-T &ConvNet& ResNet-50 &82.1  &76.13 &77.85 & 2.09 & 1.72\\
        \midrule
        SLaK-T & ConvNet&ResNet-50 & 82.5 &76.13 &\textbf{77.99}& \textbf{2.25} &\textbf{1.86}\\
        \bottomrule
		\end{tabular}		}
	\end{center}
		\vspace{-0.1in}
\end{table*}
\subsubsection{Logit-Level Distillation}

\looseness=-1 We evaluate the performance on logit-level distillation based on two  distillation methods: KD~\cite{hinton2015distilling} and NKD~\cite{yang2022rethinking}. The experiments are conducted using five teachers across convolution-based and transformer-based architectures: ViT-S~\cite{dosoViTskiy2021an}, ConvNeXt-T~\cite{liu2022convnet}, Swin-T~\cite{liu2021swin}, CSWin-T~\cite{dong2022CSWin}, SLaK-T~\cite{liu2022more}, and two student models: ConvNeXt-T~\cite{liu2022convnet} and ResNet-50~\cite{he2016deep}. We exclude ViT-S as a teacher for ConvNeXt-T since the performance of the former is worse than the latter. We follow~\cite{liu2022more} and train all models \underline{for 120 epochs} to simply show the performance trend. Later on in Section~\ref{sec:300-epochs}, we will adopt the full training recipe and train our models for 300 epochs, to enable fair comparisons with state-of-the-art models. Consequently, ``Student Top-1'' is also obtained by 120-epoch training. We sweep over temperatures $\{1, 2, 5, 10,20\}$ for all approaches. We summarize our main observations below: 

\ding{182} \textbf{Large-kernel ConvNets serve as better teachers than Transformers for small-kernel ConvNets.} Table~\ref{tab:distill_logits_120_NKD} and Table~\ref{tab:distill_120_KD} show the results of NKD and KD across various teachers, respectively. Overall, we can obverse that large-kernel teachers such as SLaK-T and ConvNeXt-T outperform all Transformer teachers by a good margin in terms of both Effective Gain and Direct Gain metrics. While the teacher accuracy of ConvNeXt falls short of the best Transformers CSWin-T by 0.6\%, its student models consistently outperform the latter by up to 0.46\%  Direct Gain and 0.55\% Effective Gain.
Moreover, Direct Gain of SLaK-T reaches 2.44\% and 1.17\% for ResNet-50 and ConvNeXt-T, respectively, which is notably higher than the best results achieved by Transformer teachers, i.e., 1.55\% and 0.65\%. 


\ding{183} \textbf{Students benefit more from larger kernels.} As teacher models, SLaK-T increases the kernel size of ConvNeXt-T from 7$\times$7 to 51$\times$51 via sparsity and factorization, outperforming the latter by 0.4\%.  Among large-kernel teachers, the student models distilled from SLaK-T consistently outperform those distilled from ConvNeXt-T under both Effective Gain and Direct Gain metrics. It indicates that the benefits of larger kernels over small kernels can be effectively distilled into student models and larger kernels teach better than the small kernels. The benefits of large-kernel ConvNets over Vision Transformers can also be generalized to lightweight ConvNets (please refer to Appendix~\ref{lightmodels}).

\ding{184} \textbf{Large-kernel ConvNets help student train faster.} 
It takes 300 epochs for ConvNeXt-T to reach 82.1\% accuracy on ImageNet under supervised training~\cite{liu2022convnet}.
It is worth noting that our distilled ConvNeXt-T, when distilled from the 51$\times$51 kernel SLaK-T, reaches the performance of its 300-epoch supervised performance in only 120 epochs, highlighting the benefits of large-to-small kernel distillation.  


\begin{table}[h]
		\vspace{-0.1 in}
	\caption{\textbf{Feature distillation with varying the number of layers $L$. Results of ResNet-50 distilled from SLaK-T and Swin-T.} Direct Gain is defined as  $acc\, (distilled) - acc\,(student)$.}
	\label{tab:distill_FD_crossLayers}
	\begin{center}
    \resizebox{0.45\textwidth}{!}{
		\begin{tabular}{c|c|cccc}
		\toprule
  \multirow{3}{*}{$L$}& No Distillation&\multicolumn{4}{c}{FD Distillation} \\
        \cline{2-6}
              & ResNet-50  &\multicolumn{2}{c}{Teacher:Swin-T}&\multicolumn{2}{c}{Teacher:SLaK-T}\\
               &Top-1  &Top-1&Direct Gain&Top-1&Direct Gain\\
        \midrule
        1 &\multirow{4}{*}{76.13} &76.77 &0.64 & 76.85 & 0.72\\
        2 &  &76.76&0.63 &76.79 & 0.66\\
        3 &  & 76.81 & 0.68 &76.94&0.81\\
        4 &  & 76.73 &0.60& 76.92 & 0.79 \\
        \bottomrule
		\end{tabular}}
	\end{center}
		\vspace{-0.1 in}
\end{table}

\subsubsection{Feature-Level Distillation}
\label{sec:feature_120}
Besides only matching the logits, many works aim to minimize the distance of intermediate features between teachers and students~\cite{heo2019comprehensive,heo2019knowledge,yang2021knowledge,huang2017like,kim2018paraphrasing}. To comprehensively understand the effect of Transformers and large-kernel ConvNets based teachers, we also conduct comparisons on feature-level distillation. Specifically, we follow the feature-level KD~\cite{wei2022contrastive} that chooses  the output of the last stage  by default to transfer knowledge. Moreover, it is common to combine logit distillation with feature distillation to further improve performance~\cite{yang2022vitkd,yang2021knowledge}. Here, we combine FD~\cite{wei2022contrastive} with NKD and evaluate different teacher models with a student model ResNet-50. Concretely, we combine feature-level and logit-level distillation by minimizing the FD loss and NKD loss together. In addition, we also conduct comparisons on multi-layer feature distillation. 
With $L=i$, it denotes that the outputs of the last $i$ number of stages are used to construct the feature distillation loss. For example, with $L=2$, the outputs of the last two stages are used for feature distillation. Results are shown in Table~\ref{tab:distill_FD} and Table~\ref{tab:distill_FD_crossLayers}.

In Table~\ref{tab:distill_FD}, we observe that large-kernel ConvNets again consistently produce more performant student models than transformer-based teachers for feature distillation. Interestingly, while SLaK-T only surpasses Swin-T by 0.08\% Direct Gain in feature distillation, the corresponding number increases to 0.66\% when incorporated with logit distillation. This result highlights the benefits of large-kernel ConvNets in combining logits and features. In Table~\ref{tab:distill_FD_crossLayers}, we observe that the student model distilled from SLaK-T outperforms those distilled from Swin-T with varying $L$, indicating the superiority of large-kernel teachers over transformer-based teachers is also held when using FD with feature maps of multiple layers, i.e. $L>1$.

\begin{table*}[t]
	\caption{\textbf{NKD~\cite{yang2022rethinking} results of ResNet-50 distilled from various teacher models on ImageNet dataset.} All models are distilled for 300 epochs. Effective Gain is defined as $\frac{acc\, (distilled) - acc\,(student)}{acc\,(teacher)}$, and Direct Gain is $acc\, (distilled) - acc\,(student)$.  }
	\label{tab:distill_300_NKD}
	\begin{center}
		\footnotesize
    \resizebox{0.9\textwidth}{!}{
		\begin{tabular}{lc|ccccccc}
			\toprule
		  Teacher &Arch. Type & Student & Teacher Top-1 &  Student Top-1  & Distilled Top-1 & Effective Gain & Direct Gain \\
            \midrule
	ViT-S & Transformer&ResNet-50 & 79.8 & 78.76 & 78.88  & 0.15 &0.12\\
      
			\midrule
		    Swin-T & Transformer&ResNet-50 & 81.3 & 78.76 & 79.44 &1.45 &1.18\\
			\midrule
	ConvNeXt-T & ConvNet&ResNet-50 &  82.1  & 
            78.76 & 80.09 &  1.6 & 1.33\\
   \midrule
            SLaK-T &ConvNet& ResNet-50 & 82.5  & 78.76 & \textbf{80.24} & \textbf{1.8} & \textbf{1.48}\\
            \bottomrule
		\end{tabular}	}	
	\end{center}
		\vspace{-0.1in}
\end{table*}

\begin{table*}[h]
\centering
\small
\vspace{-0.5em}
\caption{\textbf{Top-1 accuracy of various SOTA models on ImageNet-1K.}}
\vspace{0.5em}
\resizebox{0.9\textwidth}{!}{
\begin{tabular}{l|ccccccc}
\toprule
   Model & Method & \#Training Epochs & Image Size & \#Param. & FLOPs &  Top-1 Acc  \\
\midrule
ResNet-50~\citep{resnet} & Supervised &  90
  &   224$\times$224 & 26M & 4.1G &     76.5    \\
ResNeXt-50-32$\times$4d~\citep{xie2017aggregated}  & Supervised & 90  & 224$\times$224 & 25M & 4.3G &     77.6  \\
ResMLP-24~\citep{touvron2021resmlp} & Supervised & 400 & 224$\times$224 & 30M &  6.0G &  79.4\\
DeiT-S~\citep{touvron2021training}  & Supervised & 300 &224$\times$224 & 22M & 4.6G  & 79.8 \\ 
Swin-T~\citep{liu2021swin}  & Supervised & 300 & 224$\times$224 & 28M & 4.5G & 81.3 \\
TNT-S~\citep{han2021transformer} & Supervised & 300 &224$\times$224 & 24M & 5.2G   & 81.3 \\
T2T-ViT$_t$-14~\citep{yuan2021tokens} & Supervised & 310 & 224$\times$224  & 22M & 6.1G & 81.7 \\
ConvNeXt V1-T~\citep{liu2022convnet}  & Supervised & 300 & 224$\times$224  & 29M & 4.5G & 82.1 \\
SLaK-T~\citep{liu2022more}  & Supervised & 300  & 224$\times$224 & 30M & 5.0G  & 82.5 \\
Swin V2-T~\citep{liu2021swinv2}  & Self-Supervised & 300 &  256$\times$256  & 28M & 6.6G & 82.8 \\
ConvNeXt V2-T~\citep{woo2023ConvNeXt} & Self-Supervised & 900 & 224$\times$224  & 29M & 4.5G & 83.0 \\
ResNet-50~\cite{beyer2022knowledge} & Knowledge Distillation & 9600 & 224$\times$224  & 26M & 4.1G & 82.8 \\
\gr
ConvNeXt L2S-T   &  Knowledge Distillation & 300  & 224$\times$224  & 29M & 4.5G & \textbf{83.1} \\
\midrule
Swin-S~\citep{liu2021swin}  &  Supervised & 300  & 224$\times$224  & 50M & 8.7G & 83.0 \\
ConvNeXt V1-S~\citep{liu2022convnet}  &  Supervised & 300  & 224$\times$224  & 50M & 8.7G & 83.1 \\
SLaK-S~\citep{liu2022more} &  Supervised  & 300 &  224$\times$224 & 55M  & 9.8G & 83.8 \\
Swin V2-S~\citep{liu2021swinv2}  &  Self-Supervised & 300  & 256$\times$256  & 50M & 12.6G & 84.1  \\
\gr
ConvNeXt L2S-S  &  Knowledge Distillation & 300  & 224$\times$224  & 50M & 8.7G & \textbf{84.2} \\

 \bottomrule
\end{tabular}}
\label{tab:imagenet-1k}
\vspace{-0.1em}
\end{table*}

\begin{figure*}[h]
 \centering
\setlength\tabcolsep{1.5pt}
\settowidth\rotheadsize{Radcliffe Cam}
\resizebox{0.96\textwidth}{!}{
\begin{tabularx}{0.99\linewidth}{l ccccc }
\toprule
\rothead{\centering{NKD Method}}&\multicolumn{5}{c}{\includegraphics[width=0.9\hsize,valign=m]{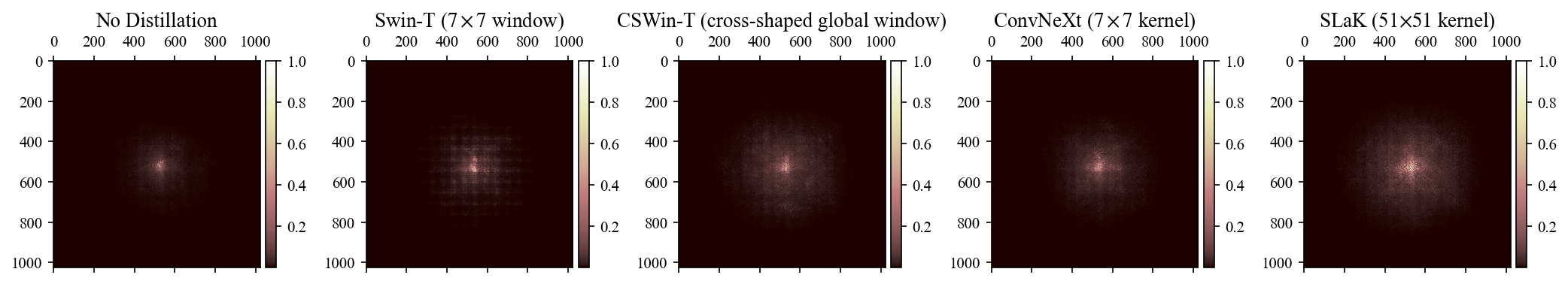}}\\[-0.0em] 
\midrule
\rothead{\centering{KD Method}}&\multicolumn{5}{c}{\includegraphics[width=0.9\hsize,valign=m]{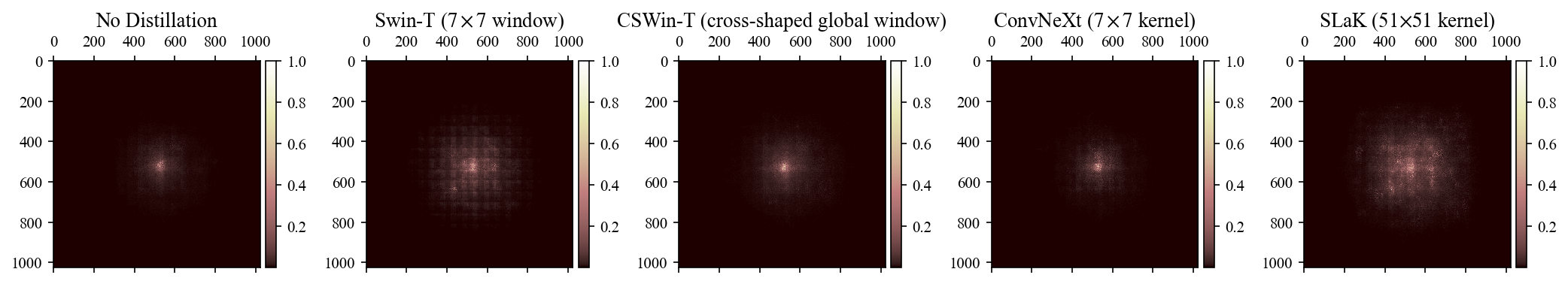}}\\[-0.0em] 
\bottomrule
\end{tabularx}}
    \caption{\textbf{Effective receptive field (ERF) of the ConvNeXt-T  distilled from various teachers}. Our student model is ConvNeXt-T with 7$\times$7 kernels. The left figures refer to the supervised ConvNeXt-T without distillation, and the rest figures are from distilled ConvNeXt-T. Overall, the students distilled from 51$\times$51 SLaK have larger and denser ERF than the students from Transformer teachers. }
\label{Fig:ERF}
\vspace{-0.5em}
\end{figure*}

\begin{table*}[htb]
\small
	\caption{\textbf{Robustness evaluation of distilled students.}}
 \vspace{ 0.5em}
	\label{tab:Robustness}
	\begin{center}
		\resizebox{0.8\textwidth}{!}{
		\begin{tabular}{lc|c|c|cccc}
			\toprule
                Teachers & Arch. Type & Student &  Clean Accuracy $\uparrow$ & C $\downarrow$ & SK $\uparrow$ & R $\uparrow$ & A $\uparrow$\\
                \midrule
                \gr
                \multicolumn{8}{c}{Robustness of Teacher Models} \\
      ViT-S&Transformer &-&79.8&47.15&26.92&39.88&12.13\\
      Swin-T&Transformer &-&81.3&46.07&29.05&41.20&21.17\\
      CSWin-T&Transformer &-&82.7&\textbf{39.47}&33.52&44.99&\textbf{31.58}\\
      ConvNeXt-T&ConvNet &-&82.1&41.99&\textbf{33.85}&\textbf{47.13}&24.02\\
      SLaK-T& ConvNet&-&82.5& 41.17&32.41&45.33&29.89\\
     \midrule
                   \gr
                \multicolumn{8}{c}{Robustness of ResNet-50 Students} \\
       -& -&ResNet-50&76.13&  54.94&26.94&35.42&4.68\\
       ViT-S&Transformer &ResNet-50&77.14&53.21&25.89&38.91&3.72\\
      Swin-T&Transformer &ResNet-50&77.67&53.46&26.07&37.59&5.45\\
      CSWin-T&Transformer &ResNet-50&77.68&53.24&27.00&38.43&6.27\\
      ConvNeXt-T&ConvNet &ResNet-50&78.04&52.43&27.74&39.02&5.91\\
      SLaK-T&ConvNet &ResNet-50&78.57&\textbf{52.09}&\textbf{28.54}&\textbf{40.16}&\textbf{7.12}\\
       \midrule
                  \gr
                \multicolumn{8}{c}{Robustness of ConvNeXt-T Students} \\
        -&-&ConvNeXt-T&81.00&44.69&32.62&45.72&20.04\\
      Swin-T&Transformer &ConvNeXt-T&81.10&44.97&31.24&43.35&17.89\\
      CSWin-T&Transformer &ConvNeXt-T&81.65&42.69&33.34&45.19&21.19\\
      ConvNeXt-T&ConvNet &ConvNeXt-T&81.77&43.17&33.82&46.49&21.06\\
      SLaK-T&ConvNet &ConvNeXt-T&82.17&\textbf{42.02}&\textbf{35.13}&\textbf{47.50}&\textbf{24.27}\\
	\bottomrule
		\end{tabular}}
	\end{center}
 	\vspace{-0.5em}
\end{table*}

\vspace{-0.5em}
\subsection{Scaling to Longer Training}

\label{sec:300-epochs}
Recent study~\cite{beyer2022knowledge} have shown that knowledge distillation requires an atypically larger number of training epochs to reach the best performance, much more than commonly used in supervised learning. Here, we also extend the training time from 120 to 300 epochs and report the performance of ResNet-50 distilled from both large-kernel teachers and transformer-based teachers. The results are shown in Table~\ref{tab:distill_300_NKD}.

We observe that the performance of ResNet-50 receives a 2.05\% performance boost when extending training epochs from 120 epochs to 300 epochs, which is in line with the observation in~\cite{beyer2022knowledge}. It is clear to see that the performance trend of longer training schedules is closely consistent with the short schedules.
The best student model is achieved by the SLaK-T teacher among all five teachers, indicating that the benefits of large-kernel teachers over transformers-based teachers also held with longer training.

\textbf{Best-Ever 30M ConvNet.} Based on the above observations, we directly distill 51$\times$51 SLaK-T into a 7$\times$7 ConvNeXt-T; SLaK-S into ConvNeXt-S, via NKD for 300 epochs, dubbed \textbf{ConvNeXt L2S-T/S}. ConvNeXt L2S-T achieves a new state-of-the-art \textbf{top-1 accuracy of 83.1\% for pure ConvNets  on ImageNet}, outperforming its supervised counterpart by 1.0\% accuracy. It is also very encouraging to observe that our distilled ConvNeXt-T, with no dedicated architectural designs, is able to slightly outperform the current state-of-the-art ConvNeXt V2-T~\cite{woo2023ConvNeXt}, which is trained by MAE-based self-supervised
learning with an improved architecture for 900 epochs in total. Our model also achieves better accuracy than the best ResNet-50 model in the literature~\cite{beyer2022knowledge} which is trained with knowledge distillation for 9600 epochs.


\vspace{-0.5em}
\section{What Else are Transferrable from Larger Kernels Teachers?}

 We then go beyond the accuracy and evaluate other important properties of students distilled from different teachers, by visualizing the effective receptive field (ERF) and testing the robustness on several ImageNet-level benchmarks. 
 
\subsection{Transferring Effective Receptive Fields (ERF)}
The concept of the effective receptive field~\cite{araujo2019computing,luo2016understanding} is an important concept in computer vision. For the output unit of one neural network layer, ERF is defined as the region containing any input pixel with a non-negligible impact on that unit~\cite{araujo2019computing}. Intuitively, anywhere in an input image outside the receptive field of a unit does not affect its output value. It is generally accepted that both large-kernel ConvNets and Vision Transformers have larger ERF, which in turn helps them outperform traditional small-kernel models.  While we have known that distillation  brings significant performance improvement to the student model, it remains very interesting to investigate whether desirable characteristics like large ERF can be distilled into small-kernel ConvNets. To do this, we visualize the ERFs of students that are distilled from large-kernel ConvNets and Vision Transformers. 

Following~\cite{ding2021repmlpnet,liu2022more}, we sample and resize 50 images from the validation set to 1024$\times$1024, and measure the contribution of each pixel on input images to the central point of the feature map generated in the last layer. The contribution scores are further accumulated and projected to a 1024$\times$1024 matrix. The visualization is shown in Figure~\ref{Fig:ERF}. We find that students distilled from SLaK-T are automatically embedded with larger and denser ERF than the ones from Swin-T and CSWin-T, albeit all of these teacher models achieve sufficient large ERF with global or large self-attention. This result also helps us better understand the success of large kernels in ConvNet distillation, that is, large kernels are more effective at distilling large REF to small kernels, improving the inherent deficiency of the latter. 

\subsection{Transferring Robustness}
Recent studies on out-of-distribution robustness~\cite{wang2022can,bai2021transformers} show that transformers and the large-kernel ConvNets embrace more robustness than the ConvNets with small-kernel. However, it is seldom explored whether such superiority in robustness can be transferred to small-kernel models through knowledge distillation. To answer this question, we directly test the different student models on several robustness benchmarks including ImageNet-R~\cite{hendrycks2021many}, ImageNet-A~\cite{hendrycks2021natural}, ImageNet-Sketch~\cite{wang2019learning}, and ImageNet-C~\cite{hendrycks2019benchmarking} datasets. Mean corruption error (mCE) is reported for ImageNet-C and top-1 accuracy is reported for the rest.

The results are shown in Table~\ref{tab:Robustness}. We can clearly see that the students distilled from modern ConvNets exhibit very promising robustness, consistently outperforming the students that are learned from state-of-the-art Transformers. Among large-kernel teachers, SLaK-T transfers better robustness to students than ConvNeXt, even though with lower robustness as teachers. However, robust Transformers do not necessarily transfer to small kernel students. For example, ResNet-50 models distilled from Swin-T and ViT-S even decrease their top-1 accuracy on ImageNet-SK/A datasets. This delivers a strong signal that large kernels are stronger teachers than advanced Vision Transformers and small kernels in terms of in-distribution and out-of-distribution. 

More comparison with state-of-the-art models can be referred to Appendix~\ref{MRE}, in which the distilled ConvNeXt student model surpasses the performance of most baseline models and attains a level of robustness comparable to the large-kernel-based model such as SLaK-Tiny, as well as high-capacity models
like Swin-B~\cite{liu2021swin} and RVT-B~\cite{mao2022towards}.

\section{Conclusions}
\looseness=-1 We have explored distilling recently popular large-kernel ConvNets into small-kernel ConvNets. This empirical study is meaningful due to the fact that small-kernel ConvNets remain valuable for practical application with limited resources.  Our paper comprehensively compares SOTA large-kernel ConvNets (ConvNeXt and SLaK) with various state-of-the-art Vision Transformers (such as ViT, Swin, and CSWin) in several KD scenarios, including logit distillation, feature distillation, transferring of ERF, and transferring of robustness. Our results demonstrate that large-kernel ConvNets are all-around stronger teachers than Vision Transformers for transferring knowledge to small-kernel ConvNets, and suggest that the merits of convolutions are not easily fading away. We hope our empirical study will further encourage the community to revisit ConvNets.

\section{Acknowledgement}
S. Liu and Z. Wang are in part supported by the NSF AI Institute for Foundations of Machine Learning (IFML). Part of this work used the Dutch national e-infrastructure with the support of the SURF Cooperative using grant no. NWO2021.060, EINF-2694 and EINF-2943/L1.

\nocite{langley00}

\bibliography{example_paper}
\bibliographystyle{icml2023}


\newpage
\appendix
\onecolumn

\section{FLOPS and Params of Teacher Models and Student Models}\label{Flops}
We carefully chose the models so that they have roughly a similar model size and FLOPs, i.e., about 5.0G FLOPs and 30M parameter count, as shown in Table~\ref{tab:params}.
\begin{table*}[h]
	\caption{\textbf{Params and FLOPS of Teacher models and Student models.}  }
        \vspace{0.1in}
	\label{tab:params}
	\begin{center}
  \resizebox{0.6\textwidth}{!}{
		\begin{tabular}{lcc|ccc}
			\toprule
		  Teacher  &Params(M)& FLOPS&Student & Params(M)&FLOPS   \\
            \midrule
			ViT-S & 22 &4.6G & ResNet-50&23&4G    \\
		    \midrule
      	Swin-T & 28&4.5G &  ResNet-50 &23&4G  \\
			\midrule
            CSWin-T &  23 &4.3G&ResNet-50 &23&4G  \\
            \midrule
		    ConvNeXt-T & 29 &4.5G&ResNet-50&23&4G \\
			\midrule
            SLaK-T   &30&5.0G& ResNet-50&23&4G  \\
			\midrule
            \midrule
		    Swin-T & 28&4.5G& ConvNeXt-T &29&4.5G  \\
            \midrule
            CSWin-T & 23&4.3G& ConvNeXt-T &29&4.5G \\
            \midrule
            ConvNeXt-T  &29&4.5G& ConvNeXt-T &29&4.5G  \\
            \midrule
            SLaK-T   &30&5.0G& ConvNeXt-T &29&4.5G  \\
            \bottomrule
		\end{tabular}}		
	\end{center}
		\vspace{-0.1in}
\end{table*}
\section{Robustness Evaluation of Distilled ConvNeXt and State-of-the-Art Baselines}\label{MRE}
We could see from Table~\ref{tab:Morerobustness} that the distilled ConvNeXt student model surpasses the performance of most baseline models and attains a level of robustness comparable to the large-kernel-based model such as SLaK-Tiny, as well as high-capacity models like Swin-B and RVT-B. This strongly suggests that distillation from large-kernel models not only enhances performance but also serves as a robustness booster. 

\begin{table}[h]
\caption[caption]{\textbf{Robustness Evaluation of ConvNeXt and Baselines}. We do not make use of any specialized modules or additional fine-tuning procedures. \label{tab:Morerobustness}}
        \vspace{0.1in}
\addtolength{\tabcolsep}{-4.5pt}
	\begin{center}
  \resizebox{0.8\textwidth}{!}{
\begin{tabular}{llcccccccc}
\toprule
        Model & Data/Size & \#Training Epoches&FLOPs / Params & Clean & C ($\downarrow$)  & A($\uparrow$) & R($\uparrow$) & SK($\uparrow$)\\
        \midrule
        RVT-S*~\cite{mao2022towards} & 1K/224$^2$ &300&  4.7 / 23.3 & 81.9 & 49.4 & 25.7 & 47.7 & 34.7  \\
        \gr
        ConvNeXt-T~\cite{liu2022convnet} & 1K/224$^2$ &300&  4.5 / 28.6 & 82.1 & 53.2 & 24.2 & 47.2 & 33.8  \\
        ConViT-S~\cite{d2021convit} & 1K/224$^2$ &300&  5.4 / 27.8 & 81.5 & 49.8 & 24.5 & 45.4 & 33.1  \\
        SLak-T~\cite{liu2022more} & 1K/224$^2$ & 300&  5.0 / 29 & 82.5 & 41.2 & 29.9 & 45.3 & 32.4  \\
        Swin-B~\cite{liu2021swin}  & 1K/224$^2$ &300&  15.4 / 87.8 & \textbf{83.4} & 54.4 & \textbf{35.8} & 46.6 & 32.4 \\
        RVT-B*~\cite{mao2022towards} & 1K/224$^2$ &300& 17.7 / 91.8 & 82.6 & 46.8  & 28.5 & 48.7 & 36.0    \\
        \midrule
        \gr
        ConVNeXt L2S-T &1K/224$^2$&120&4.5/29&82.17&42.02& 24.37&47.50&35.13\\
        ConVNeXt L2S-T & 1K/224$^2$ & 300 & 4.5/29&83.10& \textbf{40.30} &28.46 &\textbf{49.05}&\textbf{36.64}\\
        \bottomrule
    \end{tabular}
    }
    \end{center}
\vspace{-3ex}
\end{table}

\section{Experiments of ResNet-50 Student with Strong  Training Recipe from Timm}
To evaluate if large-kernel ConvNets teachers are still better than other architectures under optimal settings, we adopt the optimal training recipe of Timm and report the distilled ResNet-50 in the following table. This training recipe gives us 78.06\% top-1 accuracy with ResNet-50 on ImageNet trained for 120 epochs, which matches the Timm A1 one reported in ~\citet{wightman2021resnet}. i.e., 78.1\%. 

Results are reported in Table~\ref{tab:distill_timm_120_NKD}. Again, we see that large-kernel teachers, i.e., ConvNeXt-T and SLaK-T, consistently outperform transformer-based teachers in distilling the ResNet-50 under this optimal training recipe. We believe these results  support our claim that large-kernel ConvNets are  better teachers than Vision Transformers for small-kernel ConvNets.  

\begin{table*}[h]
	\caption{\textbf{NKD Results of ResNet-50  Distilled from Various Teachers on ImageNet.}  All models are distilled for 120 epochs and are trained with the strong training recipe from Timm.}
	\label{tab:distill_timm_120_NKD}
	\begin{center}
  \resizebox{0.95\textwidth}{!}{
		\begin{tabular}{lc|c|ccccc}
			\toprule
		  Teacher & Arch. Type & Student & Teacher Top-1 &  Student Top-1  & Distilled Top-1 & Effective Gain  & Direct Gain  \\
            \midrule
			ViT-S & Transformer & ResNet-50 & 79.8 & 78.06 & 78.50  & 0.5  &0.44  \\
		    \midrule
      	Swin-T & Transformer &  ResNet-50 & 81.3 & 78.06 & 78.76 &0.8  &0.70  \\
			\midrule
            CSWin-T & Transformer &  ResNet-50 &82.7 &78.06 & 78.80 & 0.89 & 0.74\\
            \midrule
		    ConvNeXt-T & ConvNet & ResNet-50 &  82.1  & 
            78.06 & 78.95 &  1.08 & 0.89 \\
            \midrule
            SLaK-T & ConvNet & ResNet-50 &  82.5  & 78.06&\textbf{79.14}&\textbf{1.3}&\textbf{1.1} \\
            \bottomrule
		\end{tabular}}		
	\end{center}
		\vspace{-0.1in}
\end{table*}
\section{Experiments of MobileNet Student Model}\label{lightmodels}
we conducted experiments on MobileNet-V3 and reported the results in Table~\ref{tab:distill_logits_120_NKD_light}. The results demonstrate that large-kernel models like ConvNeXt-T and SLaK-T consistently surpass all transformer-based teachers in terms of both Effective Gain and Direct Gain when paired with the lightweight model MobileNet. Our results demonstrate that the benefits of large-kernel ConvNets over Vision Transformers can be generalized to lightweight ConvNets.

\begin{table*}[h]
	\caption{\textbf{NKD results of MobileNet  distilled from various teachers on ImageNet.}  All models are distilled for 120 epochs and are trained with Timm's training recipe.}
	\label{tab:distill_logits_120_NKD_light}
	\begin{center}
  \resizebox{0.95\textwidth}{!}{
		\begin{tabular}{lc|c|ccccc}
			\toprule
		  Teacher & Arch. Type & Student & Teacher Top-1 &  Student Top-1  & Distilled Top-1 & Effective Gain  & Direct Gain  \\
            \midrule
			ViT-S & Transformer & MobileNetV3 & 79.8 &73.59  & 74.42  &1.04   & 0.83 \\
		    \midrule
      	Swin-T & Transformer & MobileNetV3 & 81.3 & 73.59 &74.53  &1.15  &0.94\\
			\midrule
            CSWin-T & Transformer & MobileNetV3 &82.7 & 73.59& 74.51 & 1.11 &0.92 \\
            \midrule
		    ConvNeXt-T & ConvNet & MobileNetV3 &  82.1  & 73.59
             &74.67  &1.31   & 1.08 \\
             \midrule
            SLaK-T & ConvNet & MobileNetV3 &  82.5  &73.59 
             &  \textbf{75.01}& \textbf{1.72}  &\textbf{1.42}  \\
            \bottomrule
		\end{tabular}}		
	\end{center}
		\vspace{-0.1in}
\end{table*}

\section{Experiments on Pre-trained Model}\label{pretraining}
We investigate the impact of knowledge distillation on the pre-trained step of a student model. 
We pre-train a student model on ImageNet-1K and fine-tune it on CIFAR-10 and CIFAR-100. We apply knowledge distillation to the ImageNet-1K pre-training step. Results are reported in Table~\ref{tab:distill_downstream_NKD} and Table~\ref{tab:distill_downstream_KD}.

The Table~\ref{tab:distill_downstream_NKD} and Table~\ref{tab:distill_downstream_KD} reveal that pre-trained student models distilled from large-kernel teachers, such as SLaK-T and ConvNeXt-T, achieve higher accuracies in downstream tasks like CIFAR-10/100 compared to those distilled from transformer-based teachers. Notably, the ResNet-50 distilled (by NKD) from SLaK-T outperforms the CSwin-T's counterpart by a significant margin (1.5\%) on CIFAR-100, highlighting its benefits. These results demonstrate the effectiveness of knowledge distillation in the pre-training stage, and moreover, large-kernel teachers provide more benefits than Vision Transformers.

\begin{table*}[htb]
	\caption{\textbf{The results of  fine-tuning the NKD distilled ResNet-50 and ConvNeXt-T models on downstream tasks CIFAR-10/100.}  All models are fine-tuned based on SGD optimizer with lr=1e-3 for 50 epochs. }
	\label{tab:distill_downstream_NKD}
	\begin{center}
  \resizebox{0.95\textwidth}{!}{
		\begin{tabular}{lc|c|ccccc}
			\toprule
		  Teacher & Arch. Type & Student  & Distilled Top-1 on ImageNet-1K & CIFAR-10 & CIFAR-100  \\
                  \midrule
      	Swin-T & Transformer &  ResNet-50 & 77.67 &96.1 & 81.6  \\
			\midrule
            CSWin-T & Transformer &  ResNet-50  & 77.68 &96.2 & 82.0 \\
            \midrule
		    ConvNeXt-T & ConvNet & ResNet-50 &  78.04 & 96.6  &82.1  \\
			\midrule
            SLaK-T  & ConvNet & ResNet-50  & \textbf{78.57} &\textbf{97.0} &\textbf{83.5}  \\
            \midrule
            \midrule
		    Swin-T & Transformer & ConvNeXt-T  &81.10 &97.4 &86.9  \\
            \midrule
            CSWin-T & Transformer & ConvNeXt-T  & 81.65 & 97.7 & 86.9\\
            \midrule
            ConvNeXt-T & ConvNet & ConvNeXt-T &81.77 & 97.9  & 87.2 \\
            \midrule
            SLaK-T & ConvNet  & ConvNeXt-T &\textbf{82.17}  &  \textbf{98.3} &\textbf{87.5}  \\
            \bottomrule
		\end{tabular}}		
	\end{center}
		\vspace{-0.1in}
\end{table*}

\begin{table*}[htb]
	\caption{\textbf{The results of  fine-tuning the KD distilled ResNet-50 and ConvNeXt-T models on downstream tasks CIFAR-10/100.}  All models are fine-tuned based on SGD optimizer with lr=1e-3 for 50 epochs. }
	\label{tab:distill_downstream_KD}
	\begin{center}
  \resizebox{0.95\textwidth}{!}{
		\begin{tabular}{lc|c|ccccc}
			\toprule
		  Teacher & Arch. Type & Student  & Distilled Top-1 on ImageNet-1K & CIFAR-10 & CIFAR-100  \\
                  \midrule
      	Swin-T & Transformer &  ResNet-50 & 76.87 &95.88 & 81.37  \\
			\midrule
            CSWin-T & Transformer &  ResNet-50  & 76.77 & 95.93&81.78  \\
            \midrule
		    ConvNeXt-T & ConvNet & ResNet-50 &  76.93 &96.15   &82.16  \\
			\midrule
            SLaK-T  & ConvNet & ResNet-50  & \textbf{77.05} &\textbf{96.34} &\textbf{82.53}  \\
            \midrule
            \midrule
		    Swin-T & Transformer & ConvNeXt-T  &80.93 & 97.17& 86.22 \\
            \midrule
            CSWin-T & Transformer & ConvNeXt-T  & 81.18 & 97.41 &86.52 \\
            \midrule
            ConvNeXt-T & ConvNet & ConvNeXt-T &81.63 & 97.69  &86.96  \\
            \midrule
            SLaK-T & ConvNet  & ConvNeXt-T &\textbf{81.86}  &  \textbf{98.02} &\textbf{87.13}  \\
            \bottomrule
		\end{tabular}}		
	\end{center}
		\vspace{-0.1in}
\end{table*}

\end{document}